\documentclass[11pt]{article}
\usepackage{coling2018}
\usepackage{times}
\usepackage{url}
\usepackage{latexsym}

\usepackage{multirow}
\usepackage{epsfig}
\usepackage{amsmath}
\usepackage{subcaption}
\usepackage{graphicx}
\usepackage{bm}

\usepackage[normalem]{ulem}
\usepackage[noend]{algpseudocode}
\usepackage{algorithm}
\usepackage{algorithmicx}

\usepackage{wrapfig}
\usepackage{lipsum}
\usepackage{epsfig}

\usepackage{amsfonts}
\usepackage{amsthm}
\usepackage{amssymb}
\usepackage{chngpage}
\usepackage{array}
\usepackage{xcolor}

\newcommand{\argmax}{\operatornamewithlimits{argmax}}

\title{Does Higher Order LSTM Have Better Accuracy for Segmenting and Labeling Sequence Data?}

\newcommand*{\affaddr}[1]{#1}

\newcommand*{\email}[1]{\texttt{#1}}

\author{Yi Zhang, Xu Sun, Shuming Ma, Yang Yang, Xuancheng Ren\\
\affaddr{MOE Key Lab of Computational Linguistics, School of EECS, Peking University}\\
\email{\{zhangyi16, xusun, shumingma, 1200012760, renxc\}@pku.edu.cn}\\
}

\date{}

\begin{document}
\maketitle
\begin{abstract}

Existing neural models usually predict the tag of the current token independent of the neighboring tags. The popular LSTM-CRF model considers the tag dependencies between every two consecutive tags. However, it is hard for existing neural models to take longer distance dependencies of tags into consideration. The scalability is mainly limited by the complex model structures and the cost of dynamic programming during training. In our work, we first design a new model called ``high order LSTM'' to predict multiple tags for the current token which contains not only the current tag but also the previous several tags. We  call the number of  tags in one prediction as ``order''. Then we propose a new method called Multi-Order BiLSTM (MO-BiLSTM) which combines low order and high order LSTMs together. MO-BiLSTM keeps the scalability to high order models with a pruning technique. We evaluate MO-BiLSTM on all-phrase chunking and NER datasets. Experiment results show that MO-BiLSTM achieves the state-of-the-art result in chunking and highly competitive results in two NER datasets. \footnote{The code
is available at \url{https://github.com/lancopku/Multi-Order-LSTM}}
  
\end{abstract}

\section{Introduction}

\blfootnote{
    \hspace{-0.65cm}  
    This work is licensed under a Creative Commons
    Attribution 4.0 International License.
    License details:
    \url{http://creativecommons.org/licenses/by/4.0/}
}

Chunking and named entity recognition are sequence labeling tasks whose target is to find the correct segments and give them the correct labels. The tags inside a segment have internal dependencies. The tags in consecutive segments may have dependencies, too. Therefore, it is natural to take the tag dependencies into consideration when making a prediction in such sequence labeling tasks.

Recently, methods have been proposed to capture tag dependencies for neural networks. \newcite{CollobertEA2011} proposed a method based on convolutional neural networks, which can use dynamic programming in training and testing stage (like a CRF layer) to capture tag dependencies. Furthermore, \newcite{huang2015bidirectional} proposed LSTM-CRF by combining LSTM and CRF for structured learning. They use a transition matrix to model the tag dependencies. A similar structure is adopted by \newcite{MaH16}. Their model also involves an external layer to extract some character level features. 

\begin{figure}[t]
	\begin{center}
		\includegraphics[width = 0.7\linewidth]{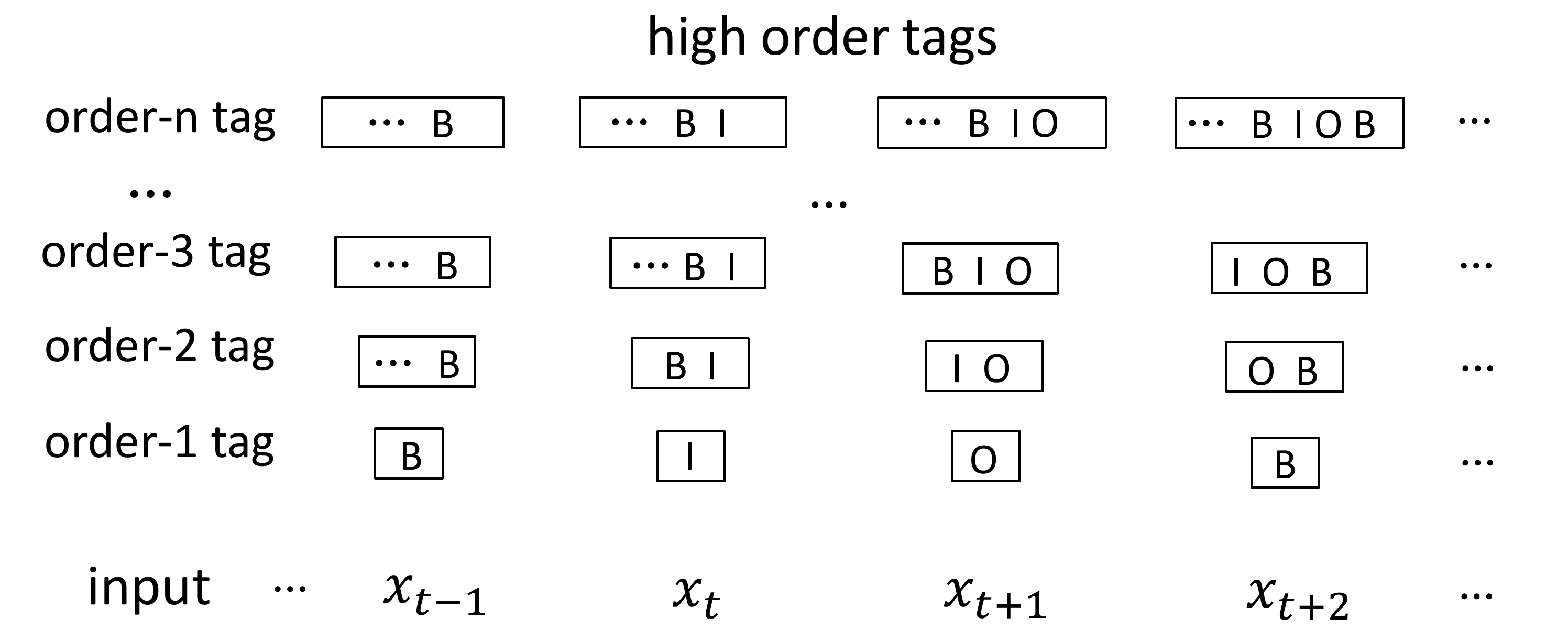}
	\end{center}
	\caption{An illustration of tags of different orders.}\label{fig1}
\end{figure}

However, it is not explicit how to model the dependencies of more tags or use the dependency information in these lines of work. We then propose a solution to capture long distance tag dependencies and use them for dependency-aware prediction of tags.
For clarity, we first give some detailed explanations of the related terms in our work. ``order'' means the number of tags that a prediction involves in a model. An order-2 tag is a bigram which contains the previous tag and the current tag at a certain time step, as shown in Figure~\ref{fig1}. Higher order tags are defined in a similar way. 

We first develop a simple method to implement high order models. But these models, which are supposed to capture more tag dependency information, perform worse and worse as the order of models increases. One possible reason is that trying to capture more tag dependencies raises the difficulty of prediction. 
We name these models as single order models and propose a new method based on them. The proposed \textbf{M}ulti-\textbf{O}rder \textbf{LSTM} (MO-LSTM) combines multi-order information from these single order models to decode. It keeps the scalability with a proposed pruning technique and performs well in our tasks. 
Experiments show that MO-LSTM achieves the state-of-the-art F1 score in all-phrase chunking and competitive scores in two NER datasets.

The contributions of this work are as follows:
\begin{itemize}
\item We extend the LSTM model to higher order models. However, the performance of the high order models which are supposed to capture longer tag dependencies is getting worse when increasing the order. 

\item We propose a model integrating low order and high order models. It keeps the scalability in both training and testing stage with a pruning technique.

\item The proposed MO-LSTM achieves an evident error reduction in chunking and NER tasks. It produces the state-of-the-art F1 score in chunking and highly competitive results in two NER datasets.
\end{itemize}

\section{Single Order LSTM}\label{single}

We first propose a simple training and decoding method which enables the existing models to extend to higher order models. Take the order-2 model as an example, for each word we combine its previous tag and its current tag to produce a bigram tag as its new tag to predict. Hence, the model can be trained with the ``new'' bigram (order-2) tag set. 

Formally, given an input sequence $\bm{x}$ = $\{x_1, x_2, \cdots, x_T\}$, where $x_t$ denotes the $t$-th word in a sentence and $T$ denotes the sentence length. The sequence $\bm{y}$ = $\{y_1, y_2, \cdots,y_T\}$ represents a possible label sequence for $\bm{x}$. We denote $\mathcal{Y}^{(1)}$ as the set of all possible order-1 labels, and $y_t \in \mathcal{Y}^{(1)}$. The order-1 model can be represented as:
\begin{equation}
    s_{1}(y_1, y_2, \cdots,y_T|\bm{x};\theta) = \prod_{t=1}^{T}s(y_{t}|\bm{x};\theta)\label{eq1}
\end{equation}
where $\theta$ is the parameters of the model. In implementation, we use a Bi-LSTM with a softmax layer to compute the score $s(y_{t}|\bm{x};\theta)$. 

To extend the order-1 model to an order-2 model, we transform the unigram label sequence into a bigram label sequence $y_0y_1, y_1y_2, \cdots,y_{T-1}y_T$, where $y_0$ is a special START symbol. The bigram label is defined as a combination of two consecutive label $y_{t-1}$ and $y_t$, and $\mathcal{Y}^{(2)}$ is the set of all possible bigram labels that appear in the training set. The order-2 model can then be written as:
\begin{equation}
    s_{2}(y_1, y_2, \cdots,y_T|\bm{x};\theta) = \prod_{t=1}^{T}s(y_{t-1}y_{t}|\bm{x};\theta)\label{eq2}
\end{equation}
Similar to the order-1 model, the score $s(y_{t-1}y_t|\bm{x};\theta)$ is computed by a Bi-LSTM with a softmax layer. In implementation, the difference with the order-1 model is that the unigram label is replaced with the bigram label. In this way, the model can be further extended to order-$n$:
\begin{equation}
    s_{n}(y_1, y_2, \cdots,y_T|\bm{x};\theta) = \prod_{t=1}^{T}s(y_{t-n+1} \cdots y_t|\bm{x};\theta)\label{eq3}
\end{equation}


As the order of the models increases, the models are supposed to learn more tag dependencies. However, according to our experiments, the performance of these models is getting worse, and the detailed results are shown in Section~\ref{exp}. An intuitive reason to explain the experimental phenomena is that the increasing size of the label set makes it more difficult to predict a correct label of the input word. Another potential reason is that the complex structure leads to overfitting problem. \newcite{DBLP:conf/nips/Sun14} suggests that complex structures are actually harmful to the generalization ability in structured prediction.

\begin{figure}[tb]
	\centering
	\subcaptionbox{Single Order-1 Model}{\includegraphics[width=0.31\linewidth]{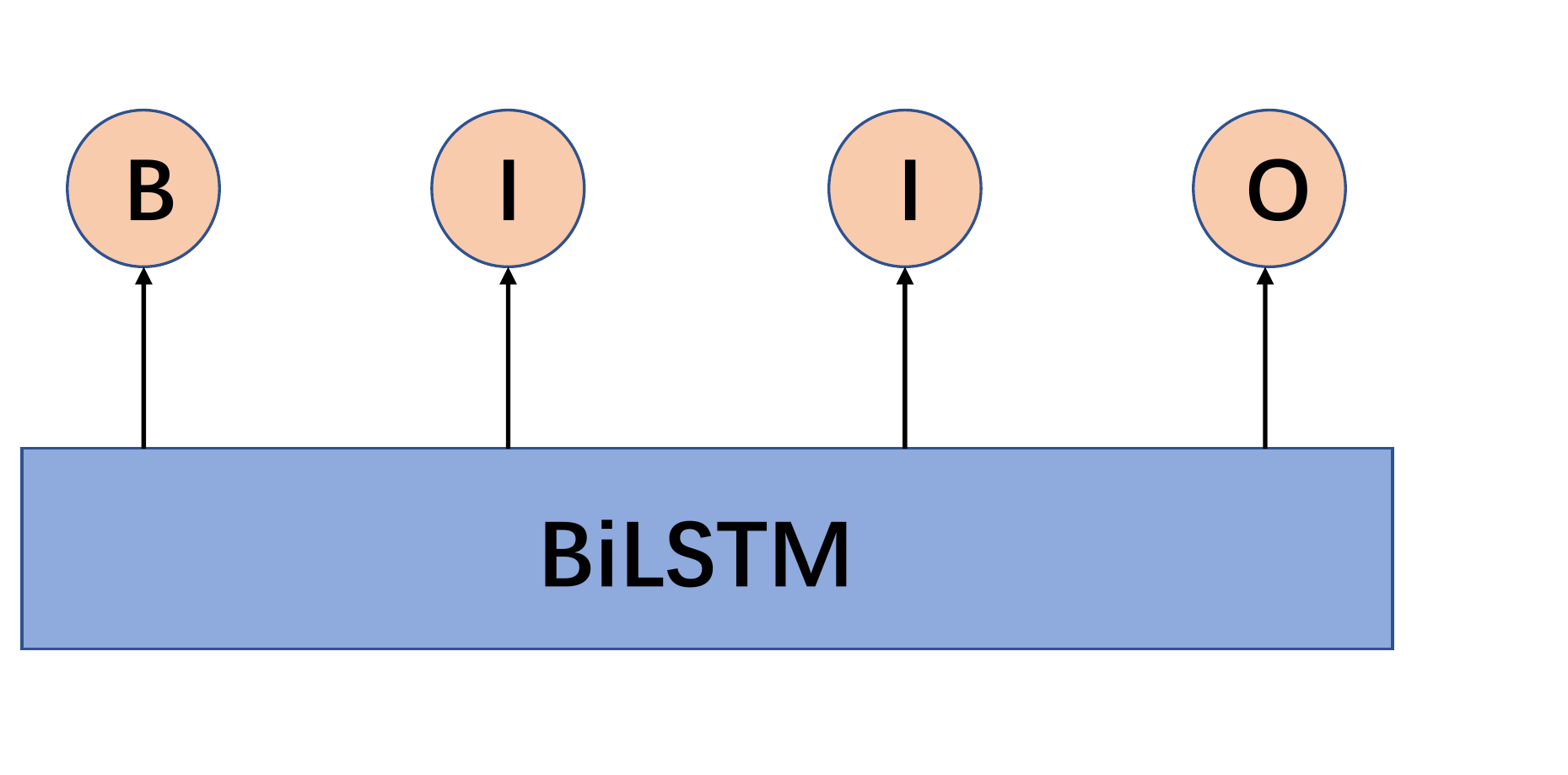}}  \quad
	\subcaptionbox{Single Order-2 Model}{\includegraphics[width=0.31\linewidth]{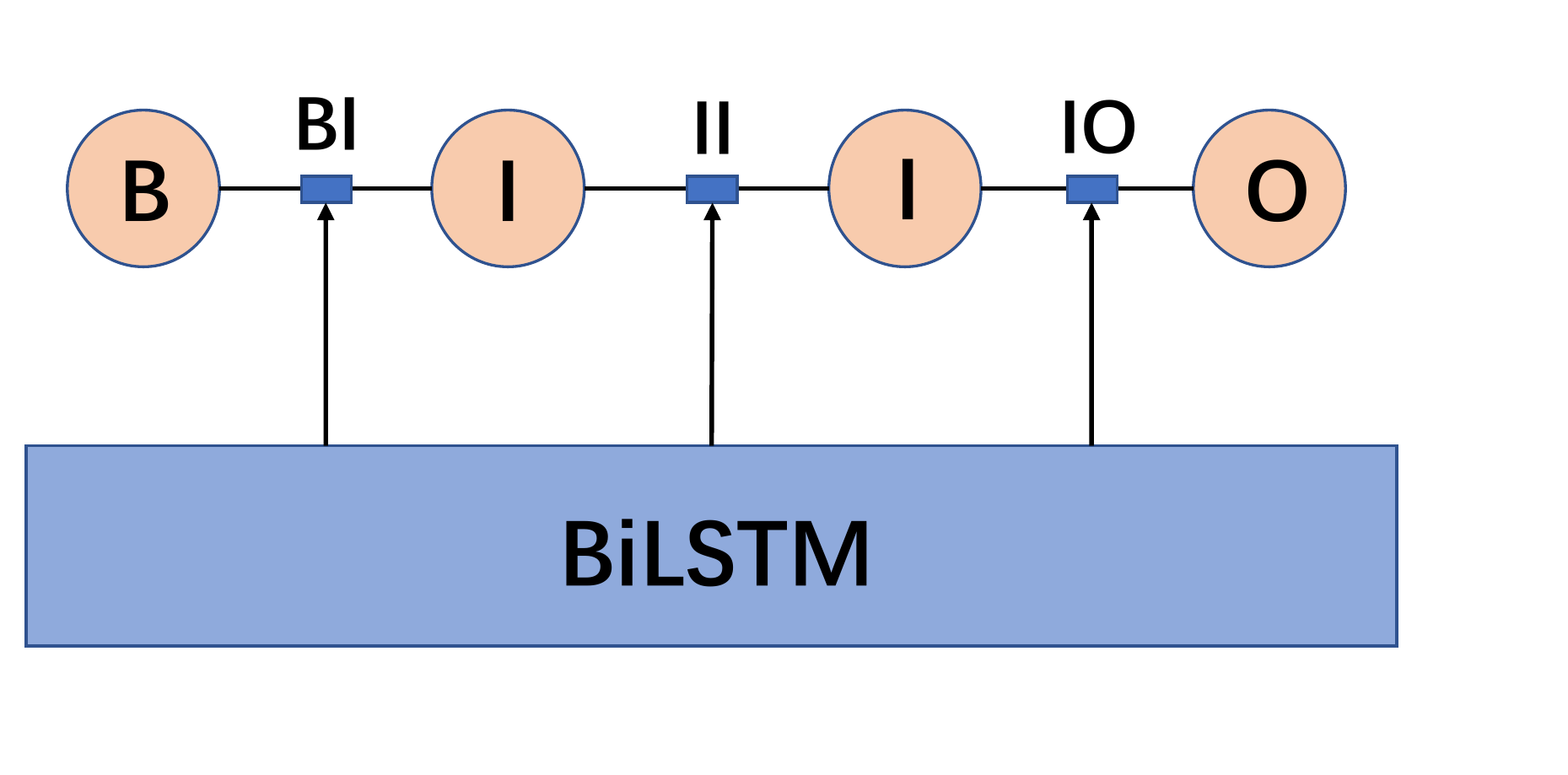}}
    \quad
	\subcaptionbox{Multi-Order-2 Model}{\includegraphics[width=0.31\linewidth]{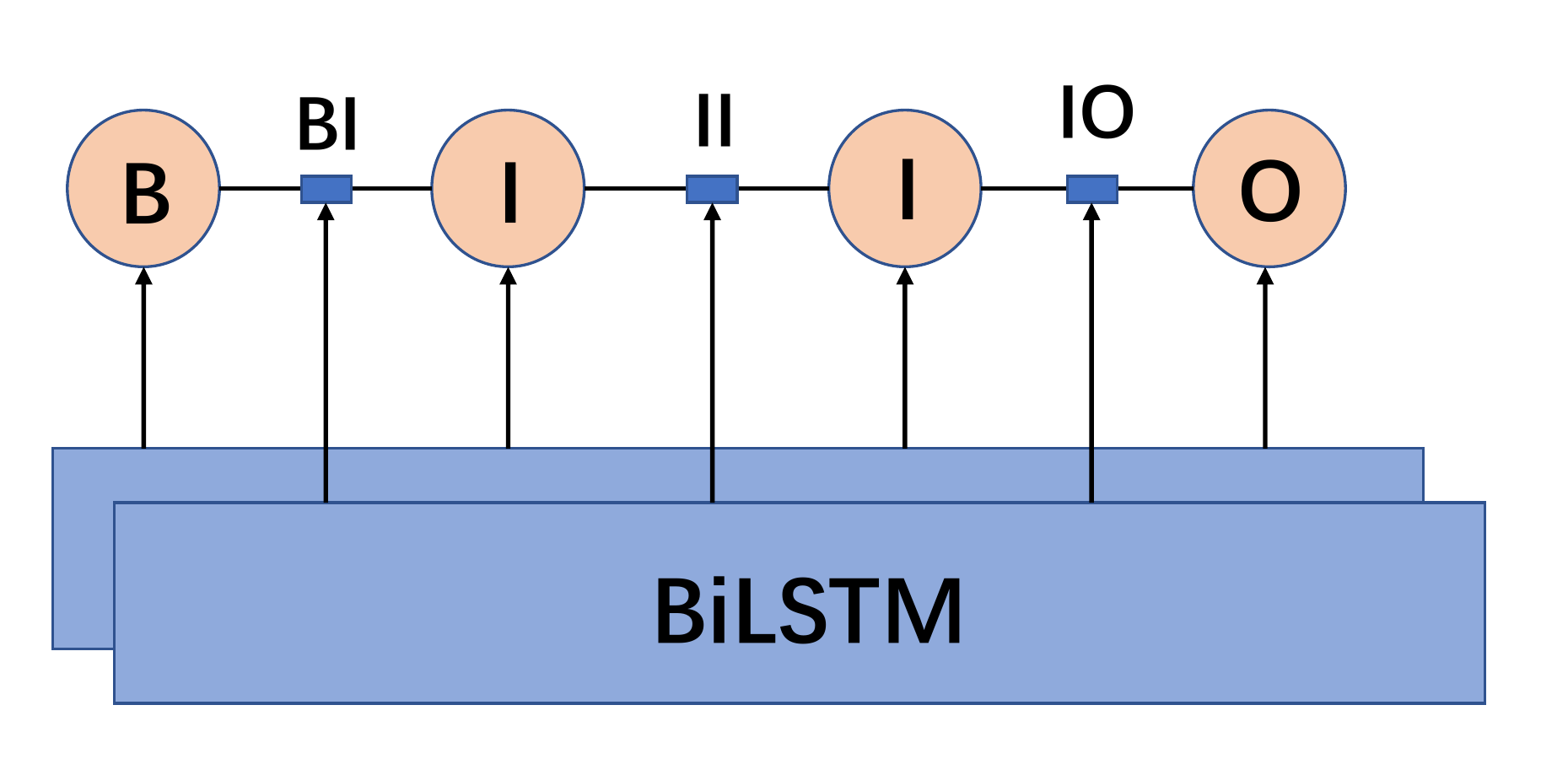}}
	\caption{An illustration of the single order model and the multi-order model. The single order-1 model is a BiLSTM. }\label{model_fig}
	\vspace{-0.05in}
\end{figure}

\section{Multi-Order BiLSTM}

The performance of single high order models deteriorates as the order increases. But they might capture some kinds of useful dependency information. To make use of these dependency information, we introduce a multi-order model which combines the low-order and high-order information. The proposed multi-order model consists of several single order models (as described in Section~\ref{single}) of different orders. At the training stage, these models are trained separately as usual. At the decoding stage, we propose a new decoding method to combine the low order model and the high order model. Since both low order information and high order information is used when decoding, the proposed method is named Multi-Order BiLSTM (MO-BiLSTM).  In this section, we first give the details of the training and the decoding process, and then introduce a pruning technique to improve the efficiency of MO-BiLSTM. 

\subsection{Multi-Order Training}

Our proposed multi-order-$n$ model is a mixture of $k$ single order models with different orders, where $n$ is the maximum order of the single order models. When $n = 1$, the multi-order model becomes a single order-1 model, i.e. a BiLSTM. The order set of the single order models is the subset of $\{1,2,\cdots,n\}$. For example, if the maximum order $n$ is 3, the combination of the single order models can be $[1,2]$, $[1,3]$, $[2,3]$, or $[1,2,3]$. Formally, we denote the order set as $\{o_1,o_2,\cdots,o_k\}$, where $o_i \textless o_j$ and $i \textless j$. In our implementation, $n$ is equal to $k$ in both training and decoding stage.

At the training stage, we train $k$ single order models separately following Eq.~\ref{eq3}:
\begin{equation}
    \theta_{i} = \argmax_{\theta}s_{o_i}(y_1, y_2, \cdots,y_T|\bm{x};\theta) = \argmax_{\theta}\prod_{t=1}^{T}s(y_{t-o_i+1} \cdots y_{t}|\bm{x};\theta)\label{eq4}
\end{equation}
where $\theta_{i}$ is the parameters of the $i$-th single order model of the order $o_i$. After training, we obtain a set of $k$ independent models: $\{s(y_{t-o_1+1} \cdots y_{t}|\bm{x};\theta_1), \cdots, s(y_{t-o_i+1} \cdots y_{t}|\bm{x};\theta_i)\}$, which learns the label dependency of different orders.

\subsection{Multi-Order Decoding}

For the purpose of simplicity and clarity, we first describe the proposed decoding method of MO-BiLSTM in the order-2 case, and then we extend it to the general order-$n$ case. 

As shown in Figure \ref{model_fig}, in the order-2 case the multi-order model is a mixture of 2 single order models, i.e. single order-1 model (Eq.~\ref{eq1}) and single order-2 model (Eq.~\ref{eq2}). At the decoding stage, the multi-order model takes account of both the order-1 model and the order-2 model. We need a new decoding approach to unify the decisions of both models. Since the order-1 model and order-2 model predict the label sequence independently, we choose to multiply the scores of order-1 model and order-2 model to get a global score, and use a dynamic programming algorithm to search for the label sequence with the maximum score:
\begin{equation}
\begin{split}
    y_1^*, y_2^*, \cdots,y_T^* &=  \argmax_{\bm{y}}s_{1}(y_1, y_2, \cdots,y_T|\bm{x};\theta_1) \times s_{2}(y_1, y_2, \cdots,y_T|\bm{x};\theta_2) \\
    &= \argmax_{\bm{y}}\prod_{t=1}^{T}s(y_{t}|\bm{x};\theta_1) \times s(y_{t-1},y_t|\bm{x};\theta_2)
\end{split}
\end{equation}
where $s(y_{t}|\bm{x};\theta_1)$ and $s(y_{t-1},y_t|\bm{x};\theta_2)$ are the score predictions of the single order-1 model and the single order-2 model, respectively. The details of the dynamic programming algorithm are shown in Section~\ref{prune}.

Further, we extend the order-2 case to a general order-$n$ case. The difference  with the order-2 case is that there are $k$ single order models to approximate the scores of the generated label sequence. We approximate the scores by multiplying all the scores of these trained single order models, and then decode the sequence with the maximum score. Formally, it can be written as:
\begin{equation}
\begin{split}
    y_1^*, y_2^*, \cdots,y_T^* &=  \argmax_{\bm{y}}\prod_{i=1}^{k}s_{o_i}(y_1, y_2, \cdots,y_T|\bm{x};\theta_i)\\
    &= \argmax_{\bm{y}}\prod_{i=1}^{k}\prod_{t=1}^{T}s(y_{t-o_i+1} \cdots y_{t}|\bm{x};\theta_i)\label{eq6}
\end{split}
\end{equation}
where $s(y_{t-o_i+1} \cdots y_{t}|\bm{x};\theta_i)$ is the score prediction of the $i$-th single order model of the order $o_i$.

\begin{algorithm*}[t]
       \caption{Multi-order decoding with pruning in the order-$n$ case}\label{alg1}
       \begin{algorithmic}[1]
       \State {\textbf{Input:} sentence $\bm{x}$, trained order-1 LSTM $s_1(y|x)$ in Eq.~\ref{eq1}, multi-order-$n$ LSTM $s_n(y|x)$ in Eq.~\ref{eq6}}
       \For {$t = 1 ...T $}    
       \State {Select the top-$k$ uni-labels by the order-1 scores:} 
       \State {\ \ \ \ $\tilde{Y_1} = {\rm topkTag}(s_1(y_t|x)), \tilde{Y_2}={\rm topkTag}(s_1(y_{t-1}|x)),\cdots, \tilde{Y_n}={\rm topkTag}(s_1(y_{t-n+1}|x))$}
       \State {Combine $n$ top-$k$ uni-label sets into a $n$-gram label set:}
       \State {\ \ \ \ $Y = \tilde{Y_1}\times \tilde{Y_2}\times \cdots \times \tilde{Y_n}$}
               \For {each $(y^1, y^2, \cdots, y^{n}) \in Y$}
              \State {Previous tag state $d_{t-1} = y^1 y^2 \cdots y^{n-1}$}
              \State {Current tag state $d_{t} = y^2y^3 \cdots y^{n}$}
              \State {Compute the transition score $s = s_n(y^1, y^2, \cdots, y^{n}|x)$} by multi-order-$n$ LSTM 
              \State {Compute the maximum score at current state $A[t][d_{t}]=max(A[t][d_{t}], A[t-1][d_{t-1}]*s)$}
               \EndFor
        \EndFor
       \State {\textbf{Output:} The optimal tag sequence $y^*$ by backtracking the path of the maximum score $A[T][d_{T}]$}
       \end{algorithmic}
 \end{algorithm*}

\subsection{Scalable Decoding with Pruning}\label{prune}
Here, we introduce an efficient dynamic programming algorithm to search for the label sequence with the maximum score. The scores of different $n$-gram labels are jointly considered in our model. Originally, we should consider all possible $n$-gram labels at every position of the sentence during dynamic programming. However, it will lead to a huge search space and a lot of time. In order to reduce the time cost, we can prune the unnecessary searching branches. For example, an order-1 model assigns a very low probability to the uni-label ``I'' of the $t$-th word, which means the order-1 model is confident that the $t$-th word can hardly be labeled as ``I''. Therefore, it is unnecessary to take account of the bi-gram labels ``I-B'', ``I-I'', and ``I-O'' at the next time step.

In implementation, we use the order-1 labels with high scores to evaluate whether to prune the high order labels. More precisely, we simply keep the top-$k$ order-1 labels at each position. The order-n labels for a specific position is generated by the top-$k$ labels of $n$ tokens around the position. Suppose a task has totally 50 labels. The order-1 model should compute 50 scores of these labels at each time step. As for the order-3 model, the number of the scores to be computed becomes $50^3$. The original search space before pruning for dynamic programming at each time step is  $50^3$. But if we only keep top-5 order-1 labels at each position and prune the order-$n$ labels, the search space will be reduced from $50^3$ to $5^3$. 

According to our experiments, the pruning technique saves a lot of time in the decoding stage and results in no loss of accuracy, and we find top-5 pruning works the best in order to balance the accuracy and the time cost. Details of the experiments can be found in Section~\ref{exp}. Algorithm ~\ref{alg1} shows the detailed process of multi-order decoding with pruning in the order-$n$ case.

\section{Experiments}\label{exp}
\subsection{Datasets}
Chunking and named entity recognition are sequence labeling tasks that are sensitive to tag dependencies. The tags inside a segment have internal dependencies. The tags in consecutive segments may have dependencies, too. Thus, we conduct experiments on the chunking and NER tasks to evaluate the proposed method. The test metric is F1-score. The chunking data is from CoNLL-2000 shared task \cite{SangBuchholz2000}, where we need to identify constituent parts of sentences (nouns, verbs, adjectives, etc.). To distinguish it from NP-chunking, it is referred to as the all-phrase chunking. We use the English NER data from the CoNLL-2003 shared task \cite{sang2003conll}. There are four types of entities to be recognized: PERSON, LOCATION, ORGANIZATION, and MISC. The other NER dataset is the Dutch-NER dataset from the shared task of CoNLL-2002. The types of entities are the same as the English NER dataset.




\subsection{Experimental Details}

Our model uses a single layer for the forward and backward LSTMs whose dimensions
are set to 200. We use the Adam learning method \cite{Adam14} with the default hyper parameters. We set the dropout \cite{JMLR:v15:srivastava14a} rate to 0.5.

Following previous work \cite{huang2015bidirectional}, we extract some spelling features and context features. We did not use extra resources, with the exception
of using Senna embeddings\footnote{Downloaded from http://ronan.collobert.com/senna/} in Chunking and English-NER tasks. The embeddings in Dutch-NER tasks are randomly initialized with a size of 50. The code is implemented with the python package \emph{Tensorflow} \cite{tensorflow}.

\begin{table*}[t] 
	\begin{center}
		\begin{tabular}{c|l|l|l}
			\hline
			\textbf{Model} & \textbf{All-Chunking} & \textbf{English-NER} & \textbf{Dutch-NER} \\
			\hline
			Single Order-1 BiLSTM &93.89 &88.23 &77.20 \\
			Single Order-2 BiLSTM &93.71 (-0.18) &87.61 (-0.62) &76.61 (-0.59) \\
			Single Order-3 BiLSTM &93.34 (-0.55) &87.47 (-0.76) &76.47 (-0.73) \\
			
            
            
            
			\hline
			
			Multi-Order-1 BiLSTM &93.89  &88.23 &77.20 \\
			Multi-Order-2 BiLSTM &94.93 (+1.04) &90.23 (+2.00)  &80.95 (+3.75)\\
			Multi-Order-3 BiLSTM &\textbf{95.01 (+1.12)}&\textbf{90.70 (+2.47)} &\textbf{81.76 (+4.56)}\\		
			
			\hline
		\end{tabular}
	\end{center}
	\caption{Results of single order models and MO-BiLSTM. The number in parentheses means the improvements or reductions compared to the results of order-1 models. All-Chunking denotes All-Phrase-Chunking.}\label{tab1}
\end{table*}

\begin{table}[t]
	\begin{center}
		\begin{tabular}{c|c|c|c}
			\hline
			\textbf{Model} & \textbf{All-Chunking} & \textbf{English-NER} & \textbf{Dutch-NER} \\
			\hline	
			Order-1 &14   &10  &11 \\
			Order-2 &154  &39  &44 \\
			Order-3 &832  &138 &158\\		
			
			\hline
		\end{tabular}
	\end{center}
	\caption{The sizes of tag set of different order.}\label{tab3}
\end{table}

\subsection{Effect of Multi-Order Setting}
For simplicity, the single order model of order-$n$ is denoted as single order-$n$ model and the multi-order model in the order-$n$ case is denoted as multi-order-$n$ model. To verify the effectiveness of MO-BiLSTM, we conduct comparison experiments of single order models and multi-order models. The results are shown in Table~\ref{tab1}. The performance of single order BiLSTM models is getting worse with the growing of the order. An intuitive reason is that the increasing size of tag set raises the difficulty to make a correct tag prediction of a word. 
Although the performance of single high order models is far from satisfactory, the multi-order models perform well with consistent growth of F1-score on three datasets. In chunking, the MO-BiLSTM at order-3 obtains a 18.3\% error reduction compared to BiLSTM. It also performs well in the NER tasks, resulting in a 21.6\% and a 20.0\% error reductions in English-NER and Dutch-NER compared to BiLSTM baselines, respectively. 

The results suggest that high order dependency information is indeed beneficial to the prediction. Furthermore, the adopted multi-order setting makes the learned tag dependency specific to the input words. The reason is that the proposed high order model encodes the tag dependency into a single ``output tag'', and model the ``output tag'' relations using a BiLSTM conditioned on the input words. The tag dependency in previous work is represented by a transition matrix, which cannot capture the relations of tag dependencies with respect to the input words. Moreover, MO-BiLSTM can take advantage of the subtle tag dependencies captured by single-order models and naturally integrate multi-order information to make tag prediction. The decoding process of MO-BiLSTM finds a global optimum tag sequence, which significantly reduces the risk of mistakes.


MO-BiLSTM also results in a growing size of tag set. The sizes of tag set from order-1 model to order-3 model are given in Table~\ref{tab3} respectively. The tag size of the model is beyond a hundred at order-3 case. Although the size of tag set grows as the order of model increases, it is acceptable in such sequence labeling problems compared to the vocabulary size in machine translation which can be over millions.

\subsection{Effect of Pruning}

The effect of pruning on speeding up the decoding is presented in Table~\ref{tab4}. As shown, the pruning technique has shown a great ability to save time with no loss of accuracy. We then give a detailed analysis of the pruning technique. Original search process of dynamic programming considers all possible high order dependencies. However, most low-order tags have been assigned very low probabilities by low-order models and they will form almost impossible high-order tags. Thus, we only keep a small subset of all low-order tags, which makes the possible combinations shrink rapidly so that the cost of dynamic programing is greatly reduced.  We also find that the pruned search space has no effect on the performance of the models.  We suppose it is almost unlikely that the best tag sequence is out of the pruned search space. Hence, the accuracy is kept to the full extent, as shown in our experiments.

\begin{table*}[h] 
	\begin{center}
		\begin{tabular}{l|c|c|c|c|c|c}
			\hline
			\multirow{2}{*}{\textbf{Model}} &\multicolumn{2}{c}{\textbf{All-Chunking}} &\multicolumn{2}{|c|}{\textbf{English-NER}}  &\multicolumn{2}{c}{\textbf{Dutch-NER}}  \\  
			\cline{2-7}
			&Time (s) &F1   &Time (s) &F1  &Time (s) &F1 \\
			\hline
			Multi-Order-2 BiLSTM w/o pruning &31.59 &94.93  &19.23 &90.23 &26.60 &80.95\\
			Multi-Order-2 BiLSTM          &13.64 &94.93  &13.13 &90.23 &18.42 &80.95\\
			\hline
			Multi-Order-3 BiLSTM w/o pruning &215.21 &95.01  &51.78 &90.70  &69.79 &81.76\\		
			Multi-Order-3 BiLSTM          &44.81  &95.01  &20.43 &90.70 &28.66 &81.76\\
			
			\hline
		\end{tabular}
	\end{center}
	\caption{Effect of pruning on speeding up the decoding.}\label{tab4}
\end{table*}

\begin{table}[t]
	\begin{center}
        \begin{tabular}{l|c}
			\hline
			\textbf{All-Chunking}  & \textbf{F1}     \\ \hline
			SVM classifier  \cite{KudoMatsumoto2001}  &93.91 \\
            Second order CRF \cite{ShaPereira2003} &94.30  \\
			Second order CRF \cite{McDonaldEA2005} &94.29 \\  
            Specialized HMM + voting scheme \cite{ShenSarkar2005} &94.01 \\
			Second order CRF \cite{Sun08} &94.34 \\
			Conv network tagger (senna)  \cite{CollobertEA2011}&94.32  \\
            CRF-ADF \cite{SunLWL14} &94.52\\
			BiLSTM-CRF (Senna) \cite{huang2015bidirectional} &94.46  \\
            Edge-based CRF \cite{DBLP:journals/corr/MaS16} &94.80\\
            Encoder-decoder-pointer framework\cite{aaai17chunking} &94.72 \\

			\hline
			BiLSTM (our implementation)  & 93.89  \\
			MO-BiLSTM (this work)  &\textbf{95.01}  \\ \hline
            
     \end{tabular}
	\end{center}
	\caption{All-Chunking: Comparison with state-of-the-art models. 
    }\label{tab6.1}
\end{table}

\begin{table}[h]
	\begin{center}
		\begin{tabular}{l|c}
			\hline
			\textbf{English-NER}  & \textbf{F1}     \\ \hline
            
			Combination of HMM, Maxent etc. \cite{florian2003named}  &88.76  \\
			Semi-supervised model combination \cite{AndoZhang05a}  &89.31 \\
            Conv-CRF (Senna + Gazetteer) \cite{CollobertEA2011} &89.59  \\
			CRF with Lexicon Infused Embeddings \cite{passos2014lexicon} &90.90  \\
			BiLSTM-CRF (Senna) \cite{huang2015bidirectional} &90.10  \\
            
			BiLSTM-CRF  \cite{LampleBSKD16}  &90.94  \\
			BiLSTM-CNNs-CRF \cite{MaH16}  &\textbf{91.21}  \\
            Iterated Dilated CNNs \cite{strubell2017fast}  &90.65 \\
            CNN-CNN-LSTM \cite{shen2018deep} &90.89 \\
			\hline
			BiLSTM (our implementation)  & 88.23  \\
			MO-BiLSTM (this work)  &90.70\\ \hline
            
\end{tabular}
	\end{center}
	\caption{English-NER: Comparison with state-of-the-art models.}\label{tab6.2}
\end{table}
\begin{table}[!h] 
	\begin{center}
		\begin{tabular}{l|c}
			\hline
			\textbf{Dutch-NER}  & \textbf{F1}     \\ \hline
			
			 AdaBoost (decision trees) \cite{carreras2002named}  &77.05 \\
			Semi-structured resources \cite{nothman2013learning}  &78.60  \\
			Variant of Seq2Seq \cite{GillickBVS15}  &78.08 \\
            Character-Level Stacked BiLSTM \cite{kuru2016charner} &79.36  \\
			BiLSTM-CRF \cite{LampleBSKD16}  &81.74  \\
            Special Decoder + Attention \cite{martins2017learning} &80.29\\ 
			
			\hline
			BiLSTM (our implementation)  & 77.20  \\
			MO-BiLSTM (this work)  &\textbf{81.76} \\ \hline
            

		\end{tabular}
	\end{center}
	\caption{Dutch-NER: Comparison with state-of-the-art models. \newcite{GillickBVS15} reported a F1-score of 82.84 in their work, but this result is based on multilingual resources. 
    }\label{tab6.3}
\end{table}
\subsection{Comparison with State-of-the-art Systems}
Table~\ref{tab6.1} shows the results on all-phrase chunking task compared with previous work. We achieve the state-of-the-art performance in all-phrase chunking. Our model outperforms the popular method BiLSTM-CRF \cite{huang2015bidirectional} by a large margin. \newcite{ShenSarkar2005} also reported a 95.23 F1-score in their paper. However, this result is based on noun phrase chunking (NP-chunking). All phrase chunking task contains much more tags to predict than NP-chunking, so it is more difficult. 

Table~\ref{tab6.2} shows the comparison results on the English-NER dataset. \newcite{MaH16} reported the best result of English NER. The main architecture of their network is BiLSTM-CRF equipped with a CNN layer to extract character-level representations of words. Our model performs slightly worse than it but outperforms BiLSTM-CRFs reported in other papers \cite{huang2015bidirectional,LampleBSKD16}.  

The comparison results on Dutch NER are shown in Table~\ref{tab6.3}. \newcite{GillickBVS15} keeps the best result of Dutch NER. However, the model is trained on four languages. With the monolingual setting, their model achieves 78.08 on F1 score. Another competitive result is reported in the work of \newcite{LampleBSKD16}. Their model is a BiLSTM-CRF model with an external LSTM layer to extract character-level representations of words. Our model gets the best score when there is no extra resources.


\subsection{Case Study}
\begin{table}[t]
\begin{adjustwidth}{0.3cm}{2cm}
 \begin{tabular}{p{.2\textwidth}| m{.7\textwidth}}

 \hline
  GOLD &  The ministry updated port conditions and shipping warnings for the \textcolor{blue}{Gulf of Mexico (LOC)}, Caribbean and Pacific Coast \\
  \hline
  BiLSTM & The ministry updated port conditions and shipping warnings for the \textcolor{magenta}{Gulf (LOC)} of \textcolor{magenta}{Mexico(LOC)} , Caribbean and Pacific Coast\\
  \hline
  MO-BiLSTM & The ministry updated port conditions and shipping warnings for the \textcolor{cyan}{Gulf of Mexico (LOC)}, Caribbean and Pacific Coast. \\
  \hline
  \hline
  GOLD & About 200 Burmese students marched briefly from troubled \textcolor{blue}{Yangon Institute of Technology (ORG)} in northern Rangoon on Friday. \\
  \hline
  BiLSTM &About 200 Burmese students marched briefly from troubled \textcolor{magenta}{Yangon (LOC)} \textcolor{magenta}{Institute of Technology (ORG)} in northern Rangoon on Friday. \\
  \hline
  MO-BiLSTM & About 200 Burmese students marched briefly from troubled \textcolor{cyan}{Yangon Institute of Technology (ORG)} in northern Rangoon on Friday.  \\
  \hline
  
 \end{tabular}
\end{adjustwidth}
\caption{Examples of the predictions of BiLSTM and MO-BiLSTM of order-3.}\label{cases}
\end{table}




We observe that MO-BiLSTM mainly helps in two aspects: the prediction of boundaries of a segment and the recognition of long segments. Table~\ref{cases} shows two cases that MO-BiLSTM model predicts correctly but BiLSTM fails to recognize the entities. In the first case, ``Gulf of Mexico'' should be recognized as the entity ``Location''. BiLSTM recognizes ``Gulf'' and ``Mexico'' as locations, but fails to recognize ``of'' as a part of the entity, so that an entire entity is split. The reason is that BiLSTM model predicts the tag independently, and it predicts ``O'' as the tag of ``of'' regardless of the neighboring tags. On the contrary, MO-BiLSTM takes account of the neighboring tags, and works well in this case. Considering that both the left tag and the right tag are labeled ``LOC'', the word ``of'' has a larger probability to be a part of the entity.

The second case contains an entity of type ``LOC''. BiLSTM succeeds in recognizing the boundary of the entity but predicts a wrong entity type for the word ``Yongon''. Although ``Yangon'' is a city, it should not be recognized as a location because it is a part of an organization. BiLSTM does not consider the neighboring tag, and makes a wrong prediction, while MO-BiLSTM succeeds in predicting a correct entity by considering the neighboring tag.

\begin{figure}[t] 

	\centering
    \subcaptionbox{Error types of the predicted entities of MO-BiLSTM.\label{pie}}{
		\includegraphics[width=0.3\linewidth]{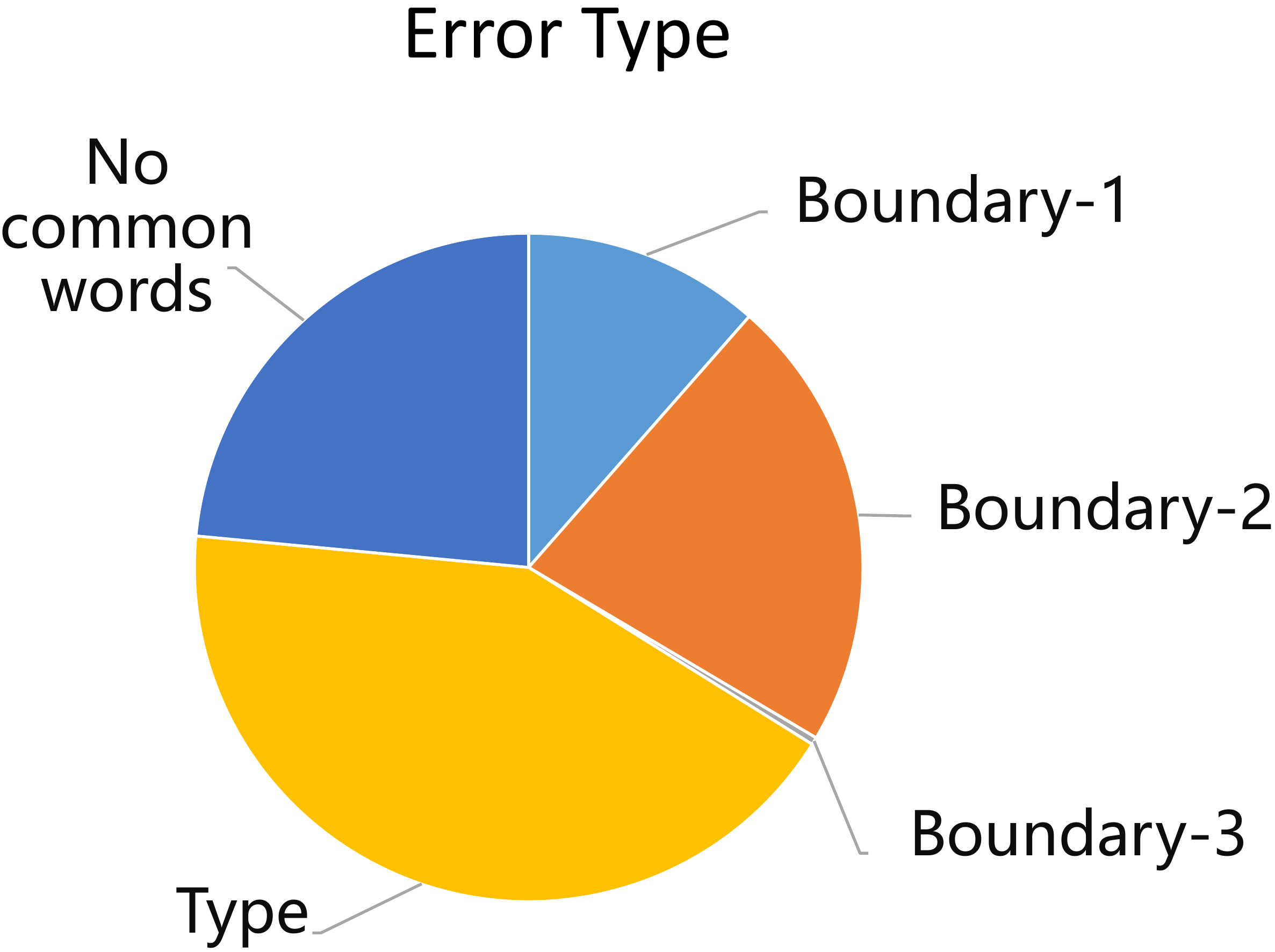}} \quad
	\subcaptionbox{Number of predicted entities belonging to boundary error.\label{boundary}}{
		\includegraphics[width=0.28\linewidth]{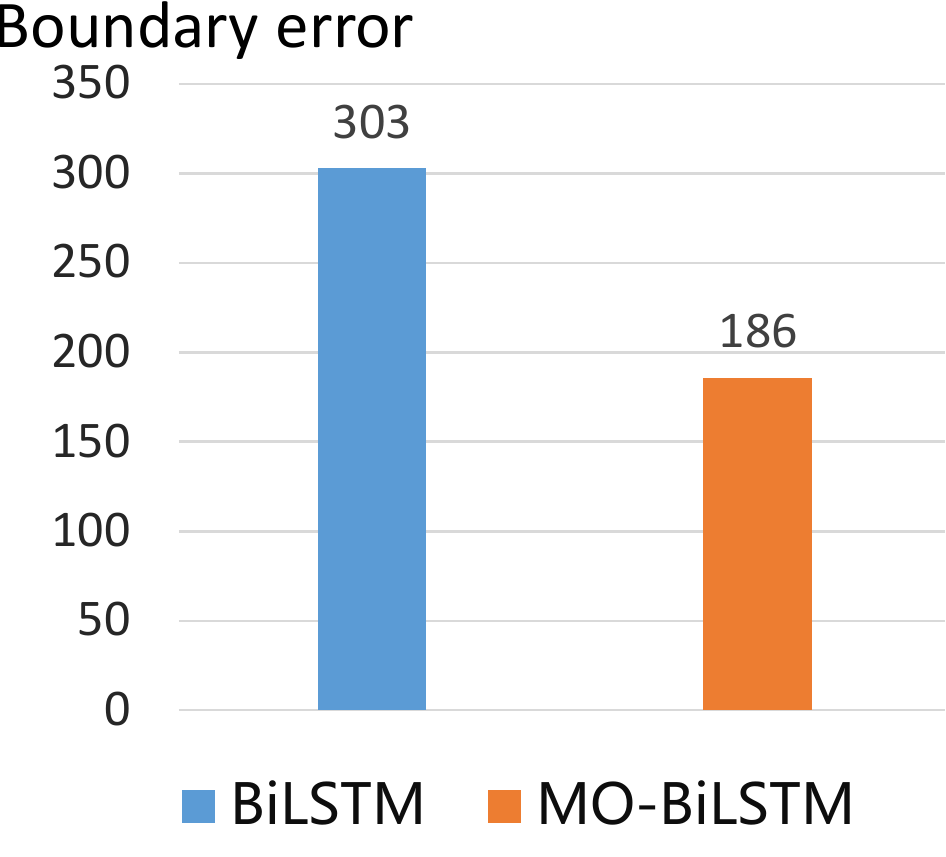}} \quad
	\subcaptionbox{Percentage of error entities regarding the length of entities.\label{length}}{
		\includegraphics[width=0.32\linewidth]{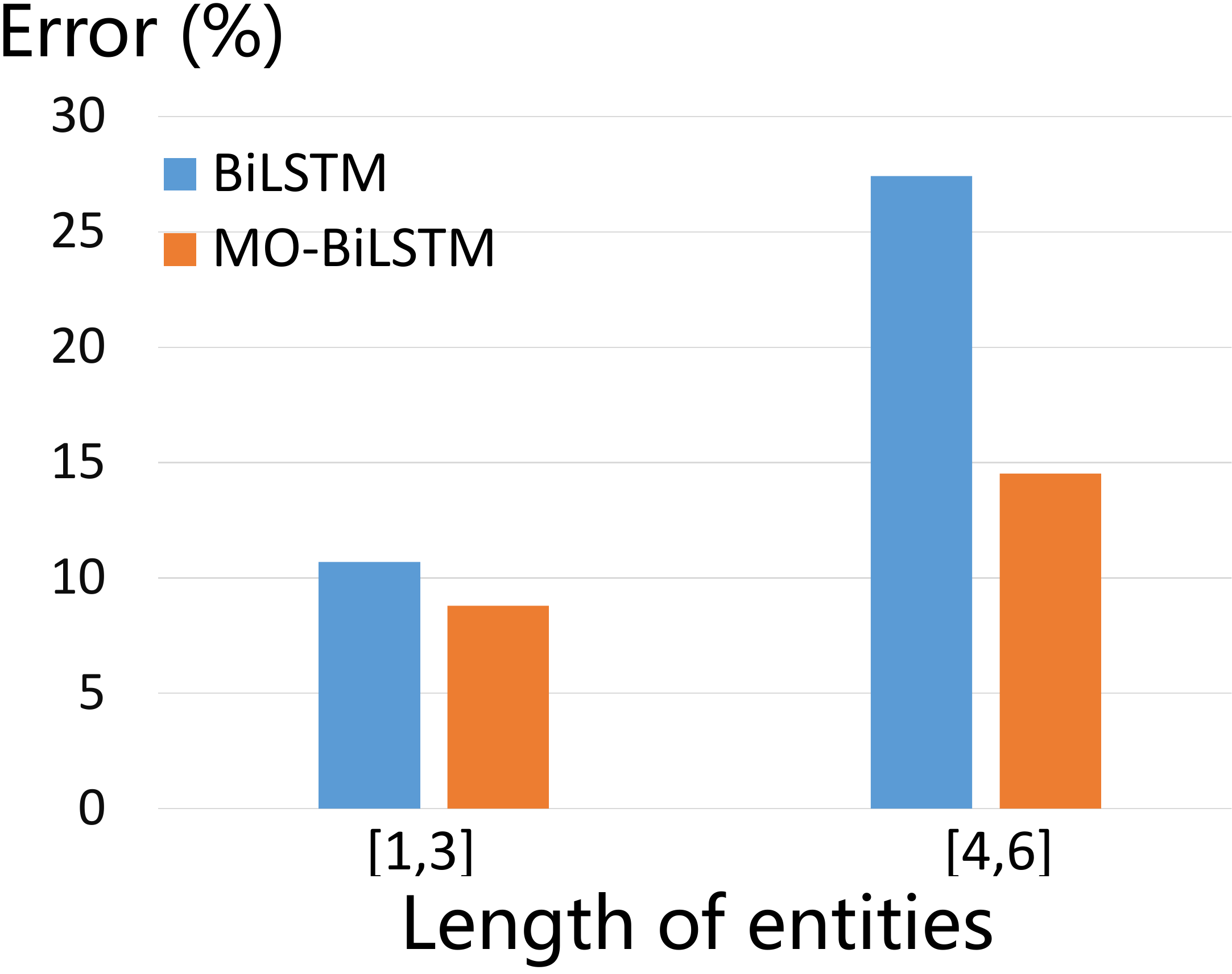}}
	\caption{Error analysis of BiLSTM and MO-BiLSTM on English-NER.}\label{bound_length}
\end{figure} 
\subsection{Error Analysis}
To better analyze the basic model and the MO-BiLSTM, we investigate the cases that can not be handled well in English-NER dataset, and the result is summarized in Figure~\ref{bound_length}. All the unrecognized entities are classified into five categories, which are ``boundary-1'', ``boundary-2'', ``boundary-3'', ``type'', and ``no common words''. ``Boundary-1'' denotes the cases that the gold entity contains a predicted entity, and ``boundary-2'' means the gold entity is contained by a prediction. ``Boundary-3'' represents the case that the gold entity and our prediction overlap. ``Type'' means a entity's boundaries are recognized correctly but its entity type is misclassified. When there are no common words between the predicted entity and any gold entity, it is denoted as ``no common words''. We count the number of wrongly predicted entities of these different categories, and the result is shown in Figure~\ref{pie}.
The ``boundary'' error (the sum of ``boundary-1'', ``boundary-2'', and ``boundary-3''), which represents the model misidentifies the entity's boundaries, is the major error type of BiLSTM. The reason is that the boundary is made up of two tags, but BiLSTM model predicts each tag independently. Our MO-BiLSTM is able to capture the dependencies between two tags, so it can significantly decrease the number of boundary recognition error. That is also the reason why ``boundary'' error is not the major error of MO-BiLSTM. 

We further compare the number of entities belonging to ``boundary'' error between BiLSTM and MO-BiLSTM. According to Figure~\ref{boundary}, it shows that the ``boundary'' error of MO-BiLSTM has a reduction rate of nearly 40\% compared with BiLSTM. 
In order to analyze the influence of the length of entities, we divide the entities into 2 groups according to their lengths, and calculate the recognition error rate of different lengths of entities. The result is shown in Figure~\ref{length}. We observe that the MO-BiLSTM model has a significant reduction in the recognition error of long entities from 27.42\% to 14.52\%. The large reduction in error rate proves that the MO-BiLSTM model is able to capture longer distance tag dependencies compared with BiLSTM.




\section{Related Work}
\newcite{huang2015bidirectional} and \newcite{LampleBSKD16} stacked a CRF layer on BiLSTM to capture the global tag dependencies. The difference between their work is the way to capture character-level information. Their proposed BiLSTM-CRF performs well in sequence labeling tasks. However, the dynamic programming must be done in both training and testing stage. The MO-BiLSTM does not need dynamic programming during training. 
 \newcite{muller2013} proposed a model that also prunes the tag set using a lower order model, but dynamic programming is required in both training and testing stage like prior work. Besides the difference that we do not need dynamic programing in training stage, the pruning technique is different. We directly model the high order states in the training stage,  while \newcite{muller2013} merges lower order states to get higher order states. \newcite{SoltaniJ16} propose a model called higher order recurrent neural networks (HORNNs). They proposed to use more memory units to keep track of more preceding RNN states, which are all recurrently fed to the hidden layers as feedback. These structures of Soltani's work are also termed ``higher order'' models, but the definition is different from ours. 
 

There are several other neural networks that use new techniques to improve sequence labeling. \newcite{LingLMAADBT15} and \newcite{YangSC16} used BiSLTM to compose character embeddings to word’s representation. \newcite{martins2017learning} used an attention mechanism to decide what is the ``best'' word to focus on next in sequence labeling tasks. \newcite{aaai17chunking} proposed to separate the segmenting and labeling in chunking. Segmentation is done by a pointer network and a decoder LSTM is used for labeling. \newcite{shen2018deep} used active learning to strategically choose most useful examples in NER datasets.

\section{Conclusions}

In this paper, we focus on extending LSTM to higher order models in order to capture more tag dependencies for segmenting and labeling sequence data. We introduce a single order model, which is supposed to capture more tag dependencies. However, the performance of the single order model is getting worse when increasing the order. To address this problem, we propose to integrate dependency information of different orders to decode. The proposed method, which is called MO-BiLSTM, keeps the scalability to high order models with a pruning technique. Experiments show that MO-BiLSTM achieves better performance than many existing popular methods. It produces the state-of-the-art result in chunking and competitive results in two NER datasets. At the end, we analyze the advantage and limitation of the MO-BiLSTM. We find that MO-BiLSTM mainly helps in the prediction of segment boundaries and the recognition of long segments.

\section*{Acknowledgements}
This work was supported in part by National Natural
Science Foundation of China (No. 61673028),
National High Technology Research and Development
Program of China (863 Program, No.
2015AA015404), and the National Thousand
Young Talents Program. Xu Sun is the corresponding
author of this paper.


\bibliographystyle{acl}
\bibliography{coling2018}

\end{document}